\documentclass[conference]{IEEEtran}
\IEEEoverridecommandlockouts
\usepackage{cite}
\usepackage{amsmath,amssymb,amsfonts}
\usepackage{hyperref}       
\usepackage{url}            
\usepackage{algorithmic}
\usepackage{graphicx}
\usepackage{textcomp}
\usepackage{xcolor}
\usepackage{balance}
\usepackage{breakurl}

\def\BibTeX{{\rm B\kern-.05em{\sc i\kern-.025em b}\kern-.08em
    T\kern-.1667em\lower.7ex\hbox{E}\kern-.125emX}}

\IEEEoverridecommandlockouts

\begin{document}

\title{Assistive Image Annotation Systems with Deep Learning and Natural Language Capabilities: A Review
}

\author{\IEEEauthorblockN{Moseli Mots'oehli}
\IEEEauthorblockA{\textit{Department of Information and Computer Science} \\
\textit{University of Hawai'i at Manoa}\\
Honolulu, USA \\
moselim@hawaii.edu}
}

\maketitle

\IEEEpubid{\makebox[\columnwidth]{979-8-3503-5326-6/24/\$31.00~\copyright~2024 IEEE \hfill} \hspace{\columnsep}\makebox[\columnwidth]{ }}
\IEEEpubidadjcol

\begin{abstract}
While supervised learning has achieved significant success in computer vision tasks, acquiring high-quality annotated data remains a bottleneck. This paper explores both scholarly and non-scholarly works in AI-assistive deep learning image annotation systems that provide textual suggestions, captions, or descriptions of the input image to the annotator. This potentially results in higher annotation efficiency and quality. Our exploration covers annotation for a range of computer vision tasks including image classification, object detection, regression, instance, semantic segmentation, and pose estimation. We review various datasets and how they contribute to the training and evaluation of AI-assistive annotation systems. We also examine methods leveraging neuro-symbolic learning, deep active learning, and self-supervised learning algorithms that enable semantic image understanding and generate free-text output. These include tasks such as image captioning, visual question answering, as well as multi-modal reasoning. Despite the promising potential, there is limited publicly available work on AI-assistive image annotation with textual output capabilities. We conclude by suggesting future research directions to advance this field, emphasizing the need for more publicly accessible datasets and collaborative efforts between academia and industry.
\end{abstract}


\section{Introduction}\label{sec:Introduction}
Deep Learning (DL) models have seen considerable success in Computer Vision (CV) tasks such as image classification, instance segmentation, Visual Question Answering (VQA), pose estimation, action recognition, and more \cite{Hans:SegmentationSurvey23,Srivastava:VQASurvey21,Zheng:PoseEstimationSurve23y}. A large proportion of this success can be attributed to the availability of large collections of annotated training data \cite{Krizhevsky:DeepCNN12}, advances in Graphics Processing Unit (GPU) technology, and advances in DL model architectures \cite{He:ResNetL16,Vaswani:RealAttention17,Kolesnikov:ViT21}. The time and financial costs associated with acquiring high-quality human annotations for large image datasets can be significant due to the dataset's scale and the need for expert annotation. Such is the case for most medical imaging datasets \cite{Bangert:ALRadiologyLabelling}, or the cost of acquisitions of the images, as is the case with satellite imagery \cite{Kremer:BigUbi17}. These challenges pose some problems in developing real-world computer vision applications that rely on DL models, thus methods are needed to reduce the error rate, time, and financial cost inherent in acquiring training annotations for DL models.

\begin{figure*}[!tbp]
\includegraphics[width=\linewidth]{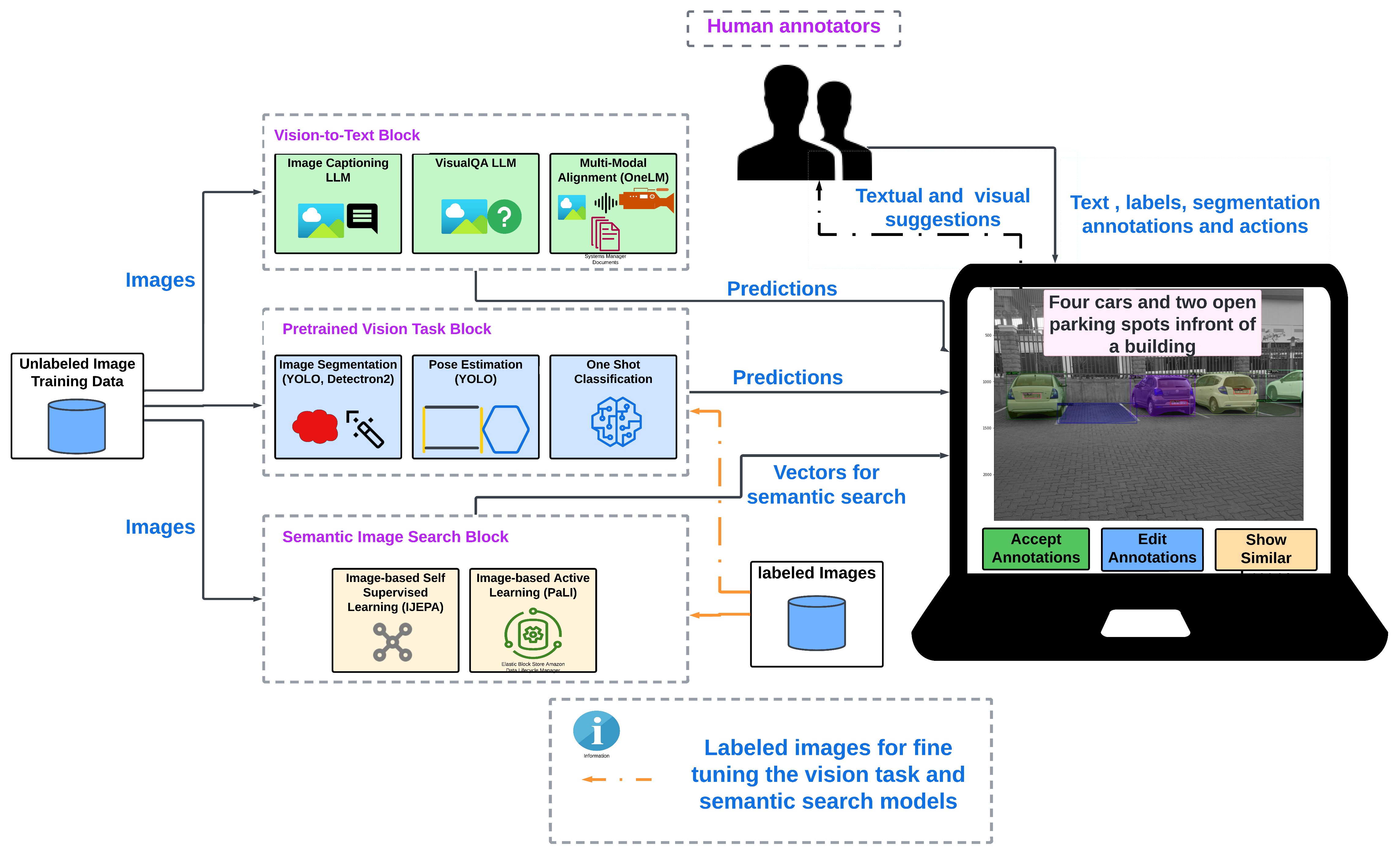}
\caption{An overview of an AI-assisted image annotation system. The system begins with unlabeled image training data which is processed through various blocks. The \textbf{Vision-to-Text Block} utilizes image captioning, VQA, and multi-modal alignment to provide predictions. The \textbf{Pretrained Vision Task Block} handles image segmentation, pose estimation, and one-shot classification to generate vision task predictions. The \textbf{Semantic Image Search Block} uses self-supervised learning and active learning to assist the annotator in semantic search. Human annotators receive textual and visual suggestions to annotate the images, which are then used to fine-tune the vision task and semantic search models. The final interface allows annotators to accept, edit, or show similar annotations.}
\label{fig:AssistiveAIImageLabelingWithTextHints}
\end{figure*}

One solution to these costs is to minimize the need for large volumes of annotated data. While techniques like self-supervised pre-training and transfer learning can help, many applications still require labeled data, especially for unique or specialized datasets. Training models on domain-specific labeled data or pre-training on related datasets can capture domain-specific nuances. and improving generalization  \cite{Chitnis:DomainSpecificPretraining23,Kataria:PretrainOrNot23}. With the advances in Self-Supervised Learning (SSL), Active Learning (AL), Few Shot Learning (FSL), and Multi-modal Learning (MML), several Artificial Intelligence (AI) assistive annotation systems have been introduced to speed up and augment the manual annotation process, reduce the cost of acquiring annotations, and improve the quality of the annotations by eliminating or flagging annotation errors as they occur, or by guiding the annotator using a combination of large language models and image understanding. These systems enhance annotation efficiency and accuracy, enabling faster DL model development and deployment in real-world applications.

To this end, this paper explores the literature on AI-assistive image annotation systems for DL models in CV tasks, with a specific interest in systems with DL or neuro-symbolic generated textual hints, descriptions, or reasoning. Textual guidance in image annotation helps annotators understand the underlying model's reasoning and focus on aspects that require human expertise, improving the overall annotation efficiency. We show an example of an AI-assistive annotation system architecture with text suggestions in Figure \ref{fig:AssistiveAIImageLabelingWithTextHints}. We compare and contrast techniques, highlighting methods from SSL, AL, FSL, and MML. Our analysis focuses on how each system addresses the challenges of annotation costs, speed, accuracy, and clarity to the annotator. We discuss evaluation metrics and benchmarks for annotation system performance, application areas, challenges, benefits, and real-world impact. We conclude this review by summarizing the state of AI-assistive image annotation systems with natural language capabilities and, explore potential avenues for future research in this area.

\section{Foundations of AI-Assisted Annotation}\label{sec:Foundations}
In this section, we briefly outline DL as it relates to computer vision tasks as well as the role of annotation in Supervised Learning (SL). We briefly provide an overview of image captioning, VQA, MML, GPT style models, and Neuro-Symbolic Learning (NeSyL). These methods share a common theme: their ability to relate images to textual descriptions. Finally, we discuss closely related surveys to this work to provide a broader context.

\subsection{Deep Learning for Computer Vision Tasks}\label{subsec:DL4CV}
Just as humans and most animals use their eyes to perceive the world for navigation and interacting with objects around them, CV is a subfield of computer science focused on creating hardware and software to assist computers in visual perception and understanding \cite{Lecun:gradientBasedLearningDocument98}. We are interested in machines with visual understanding because they can then be programmed to take actions in response to specific visual information. Many classical methods \cite{karami:SIFT2017,lorente2021image} relied on human-engineered extraction of important features. Most real-world applications of image-based Machine Learning (ML) are trained using SL, an approach to learning that, unlike unsupervised learning \cite{Barlow:UnsupervisedLearning89}, requires both the input images and the target annotations. The different CV tasks get their names from the type of their target annotations as described in Section \ref{subsec:Image_Labelling}.

Convolutional Neural Networks (CNNs)\cite{LeCun:CNN98} have been the dominant DL model architecture for learning effective image representations. They accomplish this by emphasizing the spatial relationships between neighboring pixels. By stacking convolutional layers, more complex patterns can be learned. More recently, the Vision Transformer (ViT) \cite{Kolesnikov:ViT21} architecture uses image patches and avoids convolutional layers entirely, relying solely on multi-head self-attention to process visual data.
ViTs have many training parameters, requiring substantial amounts of accurately labeled data to train without the risk of overfitting. The need for large volumes of accurately labeled data poses challenges that necessitate manual annotation, which is time-consuming, expensive, and sometimes leads to annotation errors due to factors such as fatigue, lack of focus, lack of adequate training, or just irreversible accidental mouse clicks by the annotators \cite{Northcutt:PervasiveLabelErrors21}. These errors can corrupt the training and test datasets, affecting both the training process and the reliability of the test performance evaluations.

\subsection{Image Annotation}\label{subsec:Image_Labelling}
\begin{figure}[!tbp]
\includegraphics[width=\linewidth]{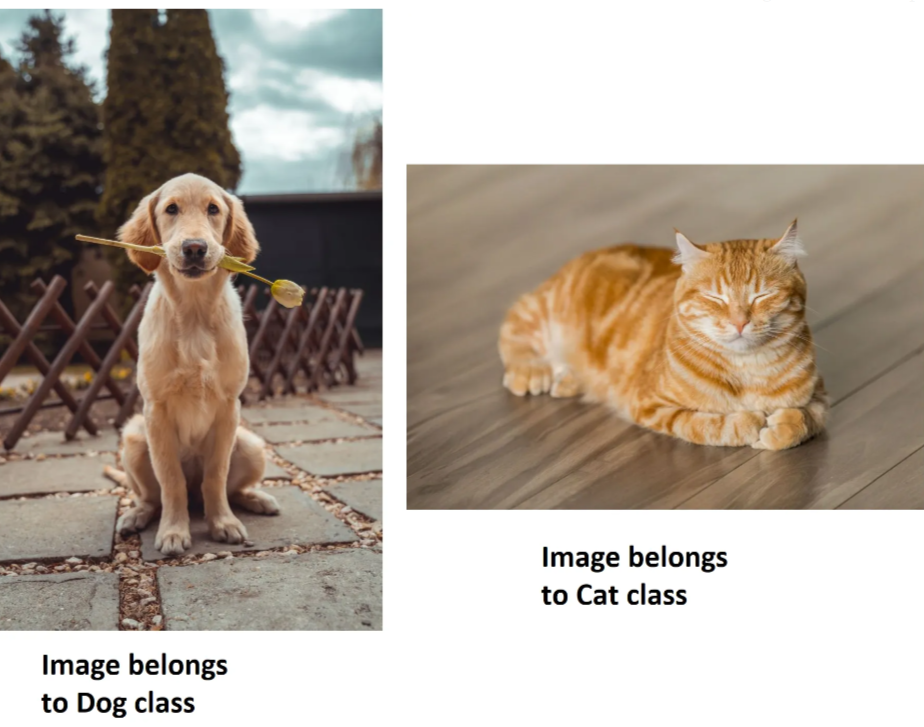}
\caption{[Source: \href{https://towardsdatascience.com/object-detection-with-convolutional-neural-networks-c9d729eedc18}{Image Classification}]: An image depicting image classification with 2 classes, a cat and a dog. The predictions are in the form of probabilities that are then mapped to the class labels based on the highest probability.}
\label{fig:classification}
\end{figure}

\begin{figure}[!tbp]
\includegraphics[width=\linewidth]{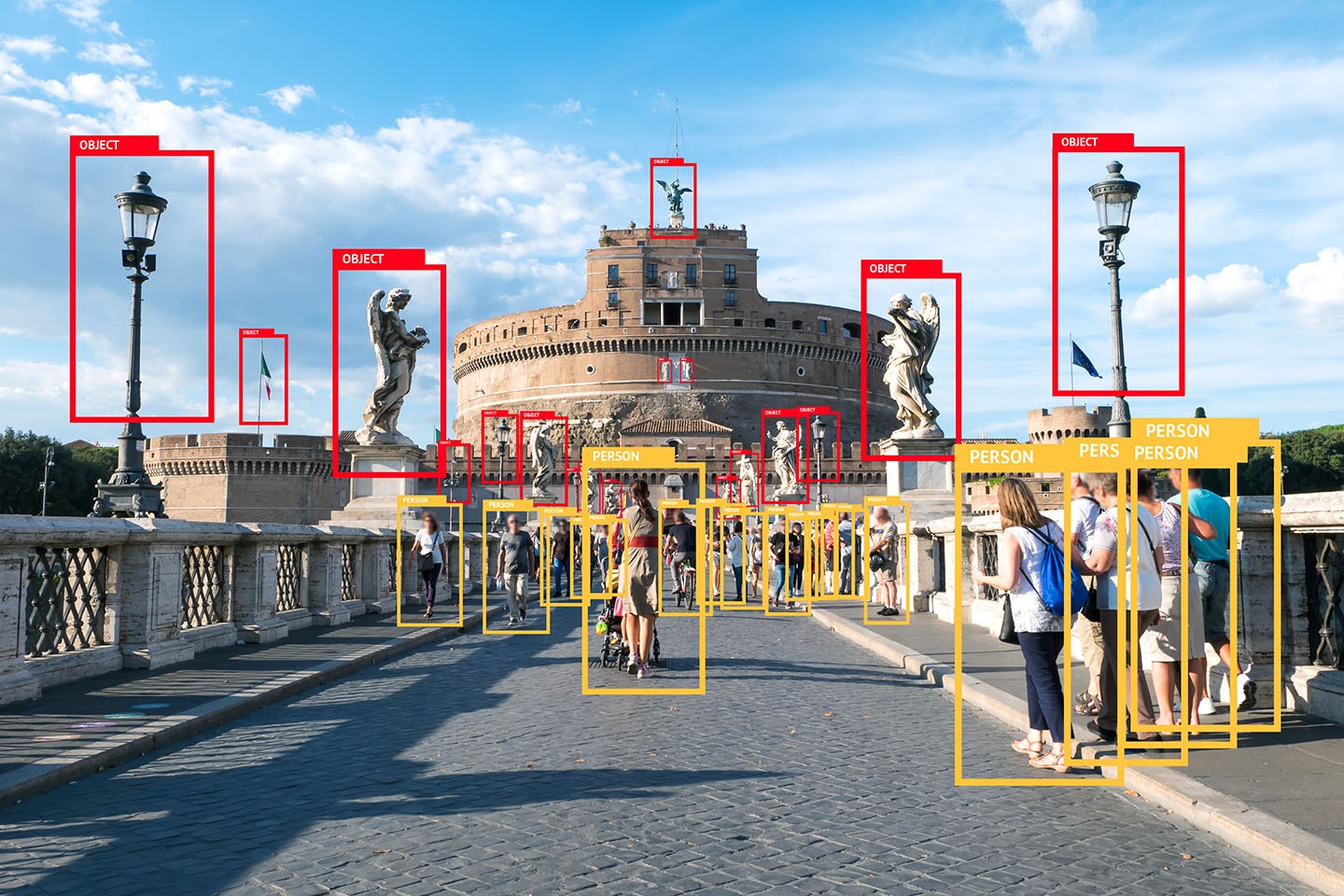}
\caption{[Source: \href{https://www.hackerearth.com/blog/developers/introduction-to-object-detection/}{Object Detection}] results on a city street showing multiple instances of humans, statues, and lights. The bounding boxes for each object are regression prediction outputs for the rectangular coordinates around each region of interest}
\label{fig:objectDetection}
\end{figure}

\begin{figure}[!tbp]
\includegraphics[width=\linewidth]{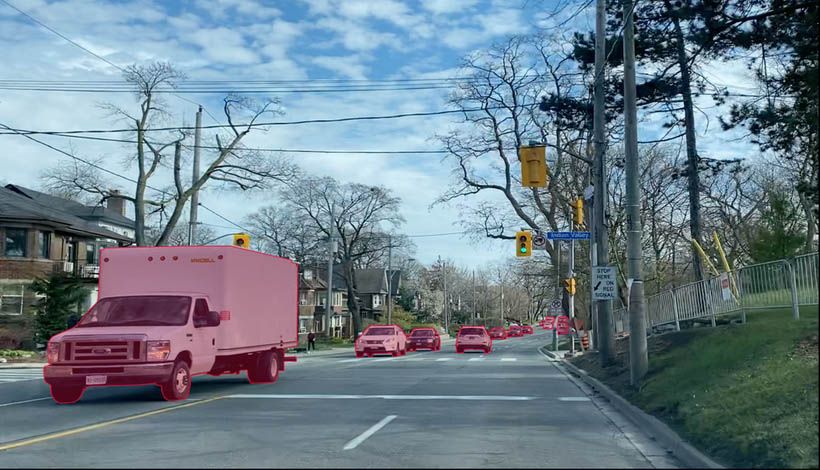}
\caption{[Source: \href{https://keymakr.com/blog/instance-vs-semantic-segmentation/}{Instance segmentation}] results of multiple cars detected and segmented at different distances from the viewpoint. Segmentation models typically provide bounding boxes and class labels for each detected object. Predictions usually consist of $K$ binary masks of size $n \times m$, outlining the pixel locations for all the  $K$ detected objects in $n \times m$ image.}
\label{fig:segmentation}
\end{figure}

\begin{figure}[!tbp]
\includegraphics[width=\linewidth]{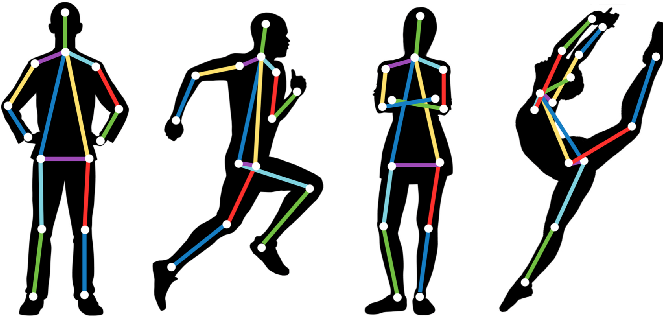}
\caption{[Source: \cite{Singh:HumanPoseEstimation19}{Example Pose Estimation}] annotations with the head, neck, shoulders, elbows, hips, knees, and ankles as key-points}
\label{fig:poseEstimation}
\end{figure}

While every CV task involves a unique annotation process with its own challenges, we limit our focus to annotation for object detection, image classification, regression, instance segmentation, and pose estimation. We outline the annotation process in each task, factors that influence the difficulty, duration, and cost of annotation, as well as potential types of errors. Figures \ref{fig:classification}, \ref{fig:objectDetection}, \ref{fig:segmentation}, and \ref{fig:poseEstimation}, display annotated images representing four of the five CV tasks (excluding regression).

\textbf{Image Classification:} Most image classification labels consist of one or more class labels for an entire image based on its perceived content \ref{fig:classification}. Labeling for classification can be negatively affected by subtle perceived differences between classes, requiring annotators to have expert-level training to accurately discriminate between them. A large number of classes can also negatively affect annotation accuracy. Labeling is generally faster and less expensive compared to other CV tasks. The most common error in classification labeling is misclassification, even in high-quality datasets such as ImageNet \cite{Recht:DoICGeneralizeImageNet19}.
\textbf{Regression:} In regression, annotators assign real continuous values to images, representing measures like length, height, or depth of the object of interest (OI). The annotation cost and complexity in regression are often dictated by the difficulty inherent in measuring the quantity of the OI, and this is normally done on the actual object and not the image. For example in fish stock estimation, the annotators are fishermen on a fishing vessel who measure the size of a live fish using a measuring board before capturing a picture of the fish and recording its species\cite{Motsoehli:fishnet24}. Annotation errors include inaccurate measurements, and typographical errors such as recording $100cm$ when the actual measurement is $10cm$. 
\textbf{Object Detection:} In object detection, the goal is to identify the presence of particular objects within an image \ref{fig:objectDetection}. During the annotation process, annotators create rectangular bounding boxes around objects of interest, and may also assign a class to each object for classification purposes. Annotation for object detection can be challenging due to factors such as very small object sizes, occlusion by other objects that are not relevant to the task, and overlapping OIs. The assignment of bounding boxes can take longer and be more expensive than annotation for classification or regression as there is a need for tight bounding boxes, that consist of four data points per object. Typical errors include inaccurate bounding boxes, mislabeled object classes, and missing bounding boxes around some objects of interest.

\textbf{Instance Segmentation:} In segmentation annotation, humans manually label each pixel in an image according to the object it belongs to. The goal is to precisely outline the boundaries of each object, enclosing it as tightly as possible. The complexity of this task varies widely based on factors such as the shape, size, and quantity of objects within the image, as well as the extent to which these OIs are obscured by background objects. Due to the increased complexity, Segmentation annotation typically takes longer and is more costly. The most common errors are missing segmentation masks for some OIs, inaccurate segmentation boundaries, or mistakenly segmenting background objects as OIs.
\textbf{Pose Estimation:} The common goal in pose estimation is detecting the position and orientation of a person or OI by identifying key points or joints \ref{fig:poseEstimation}. An example annotation involves marking and connecting the dots representing a person's head, neck, shoulders, arms, waist, knees, and feet in motion \cite{Singh:HumanPoseEstimation19}. Annotation in pose estimation is negatively affected by object occlusion, the complexity of the person's pose, image and OI size, unexpected orientational changes as well as difficulties due to a bad camera viewpoint \cite{Redondo:PoseEstimationErrors16}. The annotation process is not as complex and costly as instance segmentation annotation, but can often be more challenging than classification and regression. Common annotation errors include missing or inaccurate key points and joint markers, as well as misaligned key point markers, for instance, placing one eye key point marker slightly higher on the face than the other.

\subsection{Textual Description of Images}\label{subsec:Textual_Description_Images}
Since this work focuses on AI assistive annotation systems with natural language hint generation or reasoning, it's important to establish existing methods for learning image-to-text mappings. DL and NeSyL offer various approaches. Two prominent DL image-to-text tasks are image captioning and VQA. 
In \textbf{Image Captioning}, a CNN is typically used as an encoder, extracting a compressed representation of the image. This is then fed to a Recurrent Neural Network (RNN), typically a Long Short-Term Memory (LSTM) network \cite{Hochreiter:LSTM97}, which decodes the image representation and generates a textual description. These networks are trained jointly in an encoder-decoder fashion. \textbf{VQA systems} build upon image captioning in that the encoder receives both the image and the question text. To achieve this, the encoder combines a CNN for image representation learning and an LSTM for learning the hidden representation of the input text. The decoder, also an LSTM-like network, generates the response one token/word at a time. However, with the introduction of the ViTs and standard NLP transformer, both image and text encoders, along with the text decoder, can be built using a single architecture for image captioning and VQA \cite{Yang:AutoParsingICVQA21}.

\textbf{Multi-modal Learning (MML)} is a DL framework that bridges the gap between image and text representation learning by processing images and the accompanying text data jointly \cite{xu:MMLWithViT23}. By doing so, the model can leverage the strengths of each data modality in such a manner that the image tokens provide context for the question, and the question textual tokens guide the model to attend to specific tokens of the image in producing the response or caption. Training a joint image and text transformer architecture reduces the complexities that come with setting up and training a mix of CNN and LSTM encoders and decoders. Being able to analyze and visualize the attention maps of the Multi-modal learner can also aid in the interpretability of the question in a VQA setting or understanding of the generated textual descriptions with respect to the visual attention maps \cite{Das:MAPS22}, Current state-of-the-art MML models can handle images, audio, text, video as well as point cloud inputs \cite{zhang2023llama24,Han2023OneLLMOF24}. This can be useful in developing AI assistive image annotation systems with language generation capabilities. 

\textbf{NeSyL} image-to-text systems \cite{Wang:NeSyLCaptioning21} combine neural image representation learning with high-level symbolic representations to model image content. This is achieved by representing the objects, actions, attributes, and abstract concepts contained in an image as nodes of a knowledge graph, and the edges representing the relationships between them. For instance, an image of a boy wearing red shoes kicking a ball would have a corresponding knowledge graph with nodes "boy", "shoes", "ball", "wearing", "kicking", "red", and edges explaining the semantic, logical and spacial relations between nodes. For instance, a semantic relationship "is a" can connect two nodes "red" and "color" or "boy" and "person". This involves utilizing images, question text, and a knowledge graph as inputs to generate the expected output text. This is achieved by adjusting the DL model weights and the importance weights within the knowledge graph. Since NeSyL relies on logical rules and established domain knowledge, it minimizes the likelihood of producing text that contradicts common sense or fundamental physics laws. More recent works in NeSyL with image-to-text outputs, (Captioning, VQA, reasoning),  that are potentially applicable to AI assistive image annotation with textual hints include \cite{amizadeh:NeSyL20,Eiter:ASPNeSyL22}. 

\subsection{Related Surveys}\label{subsec:Related_Surveys}
To highlight the significance of this survey and distinguish it from prior works, we present a summary of topics covered in related surveys and identify areas that have not been addressed as they relate to this work. The most relevant survey, by Tousch et al. \cite{Tousch:SemanticHeirarchiesSurvey12}, explores prior research on semantic hierarchies for image annotation. Given that this work predates the breakthroughs in NLP and vision transformers and effective transfer learning from large language models (both appearing later in the decade), it's understandable that the authors focus on literature using structured vocabularies to describe image content for automatic annotation. Structured hierarchies of the semantics of an image are used to construct a semantic network, similar to a knowledge graph. This approach models the task of describing image contents in a NeSyL fashion. The survey is limited to annotation for CV tasks such as classification or image captioning where one or more words from a fixed structured vocabulary suffice for a valid annotation. This is not the case for image annotation with textual cues for tasks such as pose estimation, instance segmentation, and regression.     
In \cite{Sager:ImageLabelingSurveyCV21}, Sager et al. provide a comprehensive survey of image annotation for computer vision applications. By grouping annotation software into manual, semi-automated, and fully automated, show different methods used, their strength, and weaknesses. They highlight the use of clustering and transfer learning in semi-automated and fully automated annotation. However, their review does not cover the use of NLP models to generate free text and hints for AI assistive annotation that we cover in this work.

The authors of the survey \cite{Aljabri:Annotation4MedicalImagingSurvey22} follow a similar approach to \cite{Sager:ImageLabelingSurveyCV21}, but focus only on annotation software for medical imaging, specifically the graphical user interface (GUI), and component tools of the software meant to make annotation easier. Still closely related to \cite{Sager:ImageLabelingSurveyCV21}, the authors of \cite{Cheng:AutomaticImageAnnotationSurvey18} focus only on Automatic Image Annotation (AIA). Their survey groups methods into five broad categories: generative, nearest neighbors, discriminative, tag completion, and DL-based methods, based on how the annotations are automatically generated from the images. They compare the five categories of AIA based on computational complexity, time, and annotation accuracy. Similarly, this work does not address the combined use of image and text-based DL models for assistive annotation through hint generation or text descriptions. Notable surveys on Human-in-the-Loop (HITL) and human-computer joint exploration are \cite{WU:HITLSurvey22} and \cite{Wang:AssistivetaggingHumanComputerSurvey12}. The first survey \cite{WU:HITLSurvey22} reviews HITL in ML, emphasizing humans as domain experts throughout the data pipeline, from collection and annotation to model training and deployment. Our survey, however, focuses on image and text-based DL models as primary agents in annotating data for CV tasks, unlike \cite{WU:HITLSurvey22}, where humans play the central role. The second survey \cite{Wang:AssistivetaggingHumanComputerSurvey12} explores literature related to multimedia tagging. The covered methods address assistive and automated assignment, recommendation, and organization of keywords to multimedia files for internet retrieval.  However, these methods are limited to keyword selection from a dictionary and do not include free text hint generation or description in natural language leveraging DL models. Additionally, they do not cover assistive annotation in other CV tasks such as instance segmentation, pose estimation, and VQA beyond keyword tagging.

\section{Types of Assistive Deep Learning Annotation Systems}\label{sec:Assistive_Labeling_Systems}
In this section, we explore the literature on the broad systems for assistive image annotation with the help of DL, highlighting their key characteristics, limitations, and strengths. These systems leverage DL models to help human annotators in the image annotation process by generating textual hints, tags, descriptions, or logical steps. We cover Deep Active learning-based methods, self-supervised, Semi-supervised learning-based annotation systems, and Human-in-the-loop annotation platforms. We focus on the feature extraction process from input images and how it translates into textual guidance for the CV annotation task at hand.

\subsection{Deep Active Learning-based Systems}\label{subsec:Active_learning}
Deep Active Learning (DAL) seeks to train the best-performing model with as little annotated data as possible, by iteratively, and strategically selecting the most informative samples based on a DL image model, for annotation by a human annotator \cite{Ren:DALSurvey20,motsoehli2023deepActiveLearningNoisy}. In this setting, the assistive part of the annotation process is in the form of a DL model selecting the candidate images for annotation, as opposed to selecting potentially redundant samples that do not improve the quality of information in the annotated data. Most DAL methods rely on the extracted image features \cite{Rotman:MultiTaskALTransformerBased22,li:attention20}, prediction probability \cite{yang:Suggestive17}, or training dynamics \cite{wang:DALDynamics22, aklilu:ALGES22} for sample selection. Despite the extensive literature on both CV, and NLP tasks, systems leveraging AI-assisted image annotation methods that leverage DAL, and generate textual outputs remain scarce. This is likely due to limited research in this area, or the high monetization potential of such systems, resulting in proprietary industry efforts and breakthroughs remaining unpublished in scholarly articles. 

Focusing on commercial products, \cite{AmazonSageMaker:AutomatedAL} offers automated annotations based on DAL for the following CV tasks: single-label image classification, semantic segmentation, and object detection. They also handle single-label text classification. They however do not have any cross-modal, VGA, or image captioning capabilities. Roboflow \cite{Roboflow:AutomatedAnnotation}, similarly covers only a few CV tasks by leveraging large pre-trained vision models but also falls short when it comes to image-to-free-text image descriptions for annotation assistance. Labelbox \cite{Labelbox:ModelAssistedLabeling}, offers AI-Assisted auto annotation as a service covering more input modalities such as images, text, video, Geospatial data, audio, and multiple document formats. They handle most CV tasks: classification, segmentation, cloud point prediction, and object detection, as well as text-to-text annotations for NLP-type tasks. However the system is not cross-modal, it neither offers assistive annotation features for image-to-text tasks such as image captioning, and VQA, nor does it provide text hints or suggestions for image inputs that could lead to higher-quality annotations. HumanSignal \cite{HumanSignal:AutomatedAL} has increased annotator efficiency by a factor of 1.2 on the number of annotations per oracle in medical imaging through DAL-based AI-assistive annotation. Similar to \cite{Labelbox:ModelAssistedLabeling}, HumanSignal handles most input data modalities and provides assistive annotations for most CV tasks, but like SageMaker \cite{AmazonSageMaker:AutomatedAL}, HumanSignal only offers single-label image classification suggestions during the DAL cycle for assistive annotation.

HumanSignal also provides a quasi-VQA/captioning search functionality on their platform that allows the user to search the unlabeled image dataset based on predefined queries such as "find similar". While users might interpret these predefined text-based searches as a sign of natural language comprehension, the interface button most likely triggers a pre-programmed function that analyzes uploaded images and selects similar ones based on the DAL model's learned visual features. Neither the DAL implementations nor underlying vision models (CNN, ViT), for these systems are disclosed. However, assuming the class of suitable DAL algorithms at the scale these companies operate at dictates training and inference efficiency in the DAL setting as well as the interactive nature of suggestive annotation, one can rely on existing literature around the complexity and performance balance of DAL algorithms for a reasonable approximation of the underlying methods.
Telus International, CloudFactory, Encord, Datagym and Scale AI\cite{TelusInternational:AutomatedAnnotationtelus24,cloudfactory:AutomatedAnnotationcloudfactory24,Encord:AutomatedAnnotationEncord,Datagym:AutomatedAnnotationDatagym24,Scale:AutoSegmentation22} respectively cover more varied CV tasks over and above the previously stated methods at different levels of data specificity for DAL. These include keypoint pose estimations, video action recognition, autonomous driving road signs, and vehicle image mapping. However, again these systems assume adequate knowledge about the annotation task by the annotator, hence the automated assistive annotation is visual in nature, and formatted for the output of the CV task, not textual assistive hints that are capable of instilling new knowledge. While most of these tools are primarily based on DAL, they likely use pre-trained CNNs, ViTs, and rely on SSL pre-training on the input images. In the next Section, we look at methods based on SSL as well as weak supervision for AI assistive annotation with NLP capabilities.

\subsection{Self-supervised, Semi-supervised Learning-based Systems}\label{subsec:Self_supervised}
Self-supervised learning(SSL) and semi-supervised assistive image-to-text annotation systems seek to use the majority of the unlabeled data properties in both images and the accompanying text to learn meaningful image and text representations. These are normally learned by training separate Vision and Text representation models for the different input tokens in each modality. For example, in visual SSL we have generative models\cite{liu:selfGANConstrastive21}, joint embedding models \cite{chen2020simple} that predict pixel-level information, as well as joint-predictive embedding-style models \cite{assran2023selfsupervised}, that are trained to predict an intermediate representation of image patches. This forces a fundamental semantic understanding of the image contents and avoids wasted computational resources on approximating pixel-level details. In text-based foundation models, the current standard is to fine-tune an auto-regressive Large Language Model (LLM), pre-trained on next-word prediction \cite{kenton2019bert}, and some level of reinforcement learning with human feedback \cite{touvron:llama23}. Fine-tuning on a downstream task such as VQA, classification, and captioning is normally simpler for languages with large datasets, and problematic for Low-Resource-Languages \cite{ragni:data14,Marivate:PuoBerta23}. 

In AI-assistive image annotation using SSL and textual hints, the goal is to learn a good image, textual, or multi-modal representation from high-quality datasets. This is achieved by predicting parts of the input or an intermediate representation of the image, requiring only a few annotated image-text pairs to learn the mapping from image-to-text embeddings for downstream tasks. The state-of-the-art methods in this space utilize a combination of image and text SSL, before employing a fine-tuning cross-modal block that learns the image-to-text alignment \cite{Han2023OneLLMOF24}. As opposed to methods using DAL, SSL methods are normally used in the pre-training phase for representation learning. This means SSL primes suggestive DL annotation. Examples of industry-level AI-assisted annotation products based on SSL for automated image captioning and VQA include \cite{Shade:ShadeInc24,VertexAI:VertexAI,jina.ai:SSLPrompLabeling24}, all with a partial image-to-text component. Label Studio \cite{Whitaker:LabelStudio24} on the other hand focuses on LLM integration for assistive annotation for text-based inputs. They also have the same image captioning and single-target classification suggestions as the following solutions \cite{Shade:ShadeInc24,VertexAI:VertexAI,jina.ai:SSLPrompLabeling24}.
All that said, none of these methods produce free-text-type output suggestions for image annotation based on the underlying SSL model. Scholarly contributions in this area are limited but include works such as \cite{bai2019self,ferreira:UsingPseudoSSL23}, \cite{VertexAI:VertexAI}, and \cite{naithani2018automatic}. Nevertheless, these are mostly not production-level systems for image annotation, but rather experimental works on assistive annotating based on SSL. Again we see a shortage in descriptive textual interactive AI-assisted annotation methods, mainly for CV tasks such as pose estimation\cite{Neven:AcceleratingPoseLabeling23} and instance segmentation.  

\subsection{Neuro-Symbolic Learning systems}\label{subsec:Semi_supervised_and_NeSyL}
Assistive NeSyL for image annotation is a relatively unexplored area in both scholarly publications and industry applications with limited literature \cite{amizadeh:NeSyL20,Eiter:ASPNeSyL22}. This may be in part due to the current overwhelming success of DL models claiming a big share of the AI research funding and talent. It would appear NeSyL is going through it's own winter like the DL winter of the 80s. However, there are some notable NeSyL-based methods for suggestive annotation, such as those presented in \cite{Eiter:ASPNeSyL22,Adnan:SutoAnnoRetrieval15}. While NeSyL adds to the reliability and factual nature of AI systems, its strict and limited data introduces a higher annotation quality requirement in the form of domain expertise and formalism. Again, we see a deficit in free-text image annotation hints/suggestions based on NeSyL assistive annotation. 

\subsection{Human-In-The-Loop and Crowd Sourcing}\label{HITL_CrowdSourcing}
Human-In-The-Loop (HITL) image annotation systems improve AI-based auto annotation by involving a human expert annotator in the annotation process. By combining an AI-based vision model with interactive human annotation, HITL image annotation can be more efficient than manual annotation, as humans can be trained to annotate images quickly and accurately, while at the same time allowing for scalability through crowd-sourcing. The underlying AI model can be either based on SSL or a DAL model. Existing HITL and crowd-sourcing annotation systems with AI capabilities include Amazon Mechanical Turk\cite{Amazon:MechanicalTurk24} and CloudFactory\cite{Cloudfactory:AI_HITL24}. They both focus on making the annotation task affordable and fast by using multiple annotators from many places across the world and using suggestive AI models in image annotation. CrowdFlower\cite{CrowdFlower:CollectrCleanData24} focuses on providing high-quality image data annotations, metadata creation, as well as providing real-time transcriptions. The implementation details of the underlying AI models for all these commercial products are not disclosed. However, upon exploring the platforms, it's evident that none of these platforms offer assistive AI annotation tools for generating free-text outputs from images. In the next Section, we cover common evaluation metrics and benchmarks for assessing the effectiveness of assistive annotation systems.

\section{Evaluation Metrics and Benchmarks}\label{sec:Evaluations}
By using assistive annotation or automatic annotation tools, the goal is typically to improve the annotation speed, quality of the annotations, and usability of the tool. Evaluating such systems has been performed using metrics such as classification accuracy \cite{Pavoni:TagLAB21,Radeta:ManAndMachine24}, F-1 Score \cite{Hoelzemann:VideoAssistedAnnotation24,Chatterjee:AmicroN22}, Intersection over Union \cite{Pavoni:TagLAB21,VINAYAHALINGAM:DLAutoSeg23}, average annotation time \cite{VINAYAHALINGAM:DLAutoSeg23,Radeta:ManAndMachine24}, as well as metrics such as Cohen’s Kappa that measure the level of agreement between two or more annotators \cite{Hoelzemann:VideoAssistedAnnotation24}. During the evaluation, it is common to measure the perceived benefits of AI-assistive annotation for expert and non-expert annotators in the domain of the images. It has been shown that the largest improvement in annotation efficiency and accuracy due to AI-assistive annotation is seen in non-expert annotators \cite{Radeta:ManAndMachine24,Pavoni:TagLAB21}, as experts normally perform near or above the underlying DL model used in suggesting annotations. For scholarly research and model development, the following datasets in Table \ref{tab:CommonDatasets} are typically used for training and evaluating DL models capable of contributing towards assistive annotation as well as image understanding tasks. We include the dataset name, size, number of images, and the CV tasks it is capable of being used in towards AI assistive annotation.

\begin{table*}[!htp]
\begin{tabular}{lccccccccc}
\hline
 \textbf{Dataset} & \textbf{Year} & \textbf{Size} & \textbf{Detection} & \textbf{Segmentation} & \textbf{Panoptic Segmentation} & \textbf{Pose Estimation} & \textbf{Captioning} & \textbf{VQA}\\ \hline
SA-1B Dataset\cite{kirillov:SAM_1B23} & 2023 & 11M & \checkmark & \checkmark & \checkmark & \texttimes & \checkmark & \checkmark\\
VizWiz\cite{Gurari:VizWiz18} & 2018 & 20,523 & \texttimes &\texttimes & \texttimes &  \texttimes & \checkmark & \checkmark \\
CityScapes\cite{Cordts:Cityscapes16} & 2016 & ~25,000 & \texttimes & \checkmark& \checkmark & \texttimes & \texttimes & \texttimes \\
COCO-QA\cite{ren:COCO_QA15} & 2014 & 123,287 & \texttimes & \texttimes & \texttimes & \texttimes & \checkmark
& \checkmark \\
MSCOCO\cite{Lin:MSCOCO14} & 2014 & 328,000 & \checkmark & \checkmark & \checkmark & \checkmark& \checkmark & \checkmark  \\
MPII Human Pose\cite{andriluka14cvpr} & 2014 & 25,000 & \texttimes & \texttimes & \texttimes & \checkmark & \checkmark & \texttimes \\
Flickr30k\cite{young:Flickr3014} & 2014 & 31,000 & \checkmark & \texttimes & \texttimes & \texttimes & \checkmark& \texttimes \\
PASCAL VOC\cite{Everingham:Pascal10} & 2012 & 11,530 & \checkmark & \checkmark & \texttimes & \texttimes& \texttimes & \texttimes \\
\hline
\end{tabular}
\caption{Summary of common datasets: their respective capabilities for detection, instance segmentation, panoptic segmentation, pose estimation, image captioning, and visual question answering.}
\label{tab:CommonDatasets}
\end{table*}

\section{Use Cases}\label{use_Cases_Case_Studies}
AI-assistive annotation, based on underlying AI models, finds most applications in the medical imaging and biological fields due to high annotation costs, the need for expert annotators, and its potential benefits in various tasks such as instance segmentation, object detection, pose estimation, and counting  \cite{Aljabri:Annotation4MedicalImagingSurvey22,Dimitrovski:HierarchicalAnnotationMedImages11}. AI annotation also plays a role in autonomous driving, where cars equipped with sensors and high-quality cameras stream training data for crowd-sourced and AI-based annotation  \cite{swief2018survey}.
Similar patterns are observed in these areas of application. While the literature is promising in assistive annotation, image-to-textual hint suggestions remain overlooked despite the potential benefits for the annotator and performance on a predefined downstream task. 

\section{Conclusion and Future Research Directions}\label{Future_Directions_and_Conclusion}
This review has highlighted the scarcity of AI-assistive annotation systems that provide non-expert annotators with textual annotation suggestions based on image understanding and the specific CV task. We attribute the lack of progress in this area to the decades-long dependence on using different model architectures for images and text, hindering progress in cross-modal representation alignment. With advances in transformer-based representation learning \cite{caron2021emerging}, hardware acceleration, and breakthroughs in DL training, we foresee significant progress in multi-modal learning, enabling more effective and explainable AI-assistive or automated annotation. A prime example of an assistive and interactive vision-based annotation system is Meta's segment Anything model (SAM) \cite{kirillov:SAM_1B23}. SAM is capable of performing multiple CV tasks based on both visual and text-based prompts. Based on this review we see a need for developing an annotation system capable of free-text hints and suggestions for CV tasks. A viable approach would be a combination of efficient self-supervised image and language pre-training with multi-modal alignment methods and advances in NeSyL for factual grounding, as shown in Figure \ref{fig:AssistiveAIImageLabelingWithTextHints}. Additionally, promising recent image-to-text retrieval methods that may be applicable to AI-assistive annotation are presented in the following survey \cite{Wang:FullCycleCrossModalRetrieval23}.  The use of text-based annotation hints/suggestions not only lowers the requirements for expert-level human annotation but also improves the annotation speed, model interpretability, and accuracy for CV tasks. Lastly and more ambitiously, it seems obvious to us that prioritizing research around multi-modal methods capable of setting their own goals to optimize can solve most cross-domain problems. Efforts are also required to ensure such models are easily understandable, safe for use, and morally and ethically aligned with human interests, considering diverse cultures and beliefs


\section{Acknowledgments}
A big thank you to Khotso Ramoreboli and Itumeleng Tlali for the insightful discussions during the development of this paper.

\balance
\bibliographystyle{ieeetr}
\bibliography{mybibliography}

\end{document}